\documentclass[12pt]{article}
\usepackage[utf8]{inputenc}
\usepackage[T1]{fontenc}
\usepackage{amsmath}
\usepackage{amsfonts}
\usepackage{graphicx}

\usepackage{graphicx}
\usepackage{float}
\usepackage{comment}

\usepackage{tikz}
\usetikzlibrary{positioning}
\usetikzlibrary{arrows}
\usetikzlibrary{shapes.geometric}

\usepackage{amsthm}
\newtheorem*{theorem}{Theorem}

\usepackage{longtable}
\usepackage{titlesec}

\usepackage{hyperref}
\usepackage[left=3cm, right=3cm, bottom=2cm, top=1.5cm]{geometry}
\usepackage{amssymb}
\usepackage{amsmath,xspace}
\usepackage{enumitem}
\usepackage{xurl}

\title{Le Nozze di Giustizia. \\
\ \\
Interactions between Artificial Intelligence, Law, Logic, Language and Computation with some case studies in Traffic Regulations and Health Care.
}
\author{Joost J. Joosten\\
 and \\
 Manuela Montoya García}
\date{2024}

\begin{document}

\maketitle

\begin{abstract}
An important aim of this paper is to convey some basics of mathematical logic to the legal community working with Artificial Intelligence. After analysing what AI is, we decide to delimit ourselves to rule-based AI leaving Neural Networks and Machine Learning aside. Rule based AI allows for Formal methods which are described in a rudimentary form. 

We will then see how mathematical logic interacts with legal rule-based AI practice. We shall see how mathematical logic imposes limitations and complications to AI applications. We classify the limitations and interactions between mathematical logic and legal AI  in three categories: logical, computational and mathematical. 

The examples to showcase the interactions will largely come from European traffic regulations. The paper closes off with some reflections on how and where AI could be used and on basic mechanisms that shape society.
\

\medskip
\noindent\begin{keywords}\,  Artificial Intelligence, Law, Automated Decision Making, Formal Methods, Limitations to AI, Road Transportation, Mathematical Logic, Computable Laws
\end{keywords}
\end{abstract}


\section{Introduction}

What are the benefits and dangers of Artificial Intelligence (AI)? Should AI be used in the public administration and if so, how should it be used? 
What is Artificial Intelligence \emph{exactly}? What can and what can not be done with AI? 

These are main and important questions that shall be addressed in this paper. This paper is aimed primarily at the scholar with little to no background in computer science or formal methods. We mainly think of the legal scholar here or someone from the political sciences. 

As such from an educational perspective we will deal with some basics from theoretical computer science. These basics will clearly show us some limits of artificial intelligence. Sometimes one can read outlandish claims in the press on AI. With a minimal knowledge of Logic and Theoretical Computer Science, one can see that various claims on AI can just be put aside as nonsense.

In this paper we will thus be concerned at discussing limitations of AI. In Section \ref{section:WhatIsAI} we discuss what exactly AI is and delimit our domain of discourse to a particular kind of AI. The limitations will then be presented into three kinds of categories. The first category is logic and is dealt with in Section \ref{section:LogicalLingerings}: due to mere logic, we shall see that certain computational tasks are impossible. They are impossible with or without AI, simply impossible in the virtue of logic. 

Next, in Section \ref{section:ComputationalComplications} we will see a second category of limitations to AI that we call \emph{Computational Complications}. We will see that certain tasks are in principle easy to compute. So they are not subject to the logical impossibilities as from the category of logic. However, even though the computational tasks we describe are easy to solve \emph{conceptually speaking}, it may require more computational resources (computational time and memory) than available in the visible universe. 

In Section \ref{section:MultipleMathMischief} we will collect some remaining issues under the name of \emph{Mathematical Mischiefs}. In particular, we will see how easy it is to identify legal stipulations that give rise to wild mathematical behaviour; Behaviour that the writer of the law most likely simply was not aware of, but that naturally manifests itself in a mathematical automated environment and in particular in the case of \emph{computable laws}.

\section{What is Artificial Intelligence?}\label{section:WhatIsAI}

In this section we look at the nature of Artificial Intelligence and consider if it should be used.

\subsection{What is intelligence?}

Before answering the question, \emph{what is Artificial Intelligence}, let us focus on the word  \emph{Intelligence}. So we first pose the question: what is intelligence? 

As with most substantial concepts, there is no clear-cut definition for it. In particular, there are different schools and opinions on what intelligence exactly is. The Boring answer is to say
\begin{quote}
Intelligence is what intelligence tests measure.
\end{quote}
This is a pragmatic turn taken by Edwin Boring in his influential paper in 1923 on an intelligence test \cite{Boring:Intelligence:1923}. Boring is seen as a pioneer in Cognitive Psychology and there are still people today that defend his definition. In this context, it is then curious to observe that spring 2023 ChatGPT obtained an IQ score of 155 on the verbal part of the IQ test \cite{Roivainen:IQChatGPT:2023}. ChatGPT could only be tested on skills that can be plugged into the IQ test. For example, spatial puzzles and image processing can at the current date not yet be fed to ChatGPT. But still, an IQ of 155 for those skills that only require language interaction is impressive. 

Should we thus conclude that ChatGPT is intelligent? We doubt it. However, there are various other views on intelligence. Rather than exposing the various definitions and views, we would instead just mention some attributes that in most of the accounts can be assigned to intelligent behaviour. As such we can remain agnostic on the precise definition which we consider a rather honest viewpoint.  In the following paragraph we will briefly mention some of the more common features that are usually attributed to Intelligent behaviour and we shall highlight those features through the use of a cursive font. 

Intelligent behaviour should certainly include \emph{linguistic abilities} and somehow should be related to the ability to \emph{think} or \emph{reason correctly}. In particular, intelligent behaviour should have \emph{predictive power} and be \emph{adaptive to the environment} and as such it should be able to \emph{adequately change} \emph{responses, perspectives, predictions} and \emph{judgements} if the changing environment so requires. In particular, intelligent behaviour should comprise \emph{abstract thinking}, \emph{conjecturing hypotheses} and \emph{computation}. Real intelligence would combine all these attributes in an \emph{ample context} that takes \emph{common knowledge} into account and real intelligence should include \emph{compasion} and the \emph{human dimension}.

As we said, we will not pick a particular definition of intelligence to work with. Rather, we trust that we recognise intelligent behaviour when confronted with it. This is rather similar to the way we deal with concepts of time or number. Full understanding of time and number are not needed to work with them.

\subsection{Then, what is artificial intelligence?}\label{WhatIsAI:subsectionThenArtificial}

In the previous subsection we dwelled on the question what intelligence is. Now let us address what \emph{artificial intelligence} is. First we ask ourselves the question: is the word \emph{artificial} in \emph{artificial intelligence} really an informative modifier on the concept of \emph{intelligence}? Linguists (e.g.~\cite{JespersenCarara:Malfunction:2011}) would phrase this question as whether \emph{artificial} is a subsective modifier of the noun \emph{intelligence} or does the full phrase \emph{artificial intelligence} rather behave as a privative?
A privative is like a new denominator that is not compositional, just as the animal \emph{flying fox} refers to a bat and has nothing to do with foxes. 

In the light of this question it is good to mention an interesting observation made in the classical textbook on Artificial Intelligence \cite{RussellNorvig:AIAmodernApproach:2010}. They elaborate on an analogue to flying. For a long time, humanity tried to fly. Early attempts were all based on imitating flying creatures in nature. Only when inventors disconnected from the natural mechanisms of flying creatures, mankind was able to fly using the technique of a propellor which does not occur as such in nature. In analogy, working artificial intelligence could well be very different of nature as human intelligence.

At the beginning of the AI era, it was mainly logical methods that were used to mimic human reasoning. Recently we have seen huge advances with Large Language Models and Neural Networks where the latter are deterministically working reinforced networks which are clearly inspired by natural neural networks. As far as logical reasoning is concerned, one may ask ourselves the question to what extent humans take decisions and set out strategies based on thorough logical reasoning. In a by now famous experiment, P. C. Wason showed (\cite{Wason:ReasoningAboutARule:1968, WasonShapiro:Reasoning:1971}) that humans do not really follow logical rules when reasoning about implications and other logical constructs. It thus seems that logic may not be the main motor of human thinking but that logic is only resorted to after a strong intellectual effort looking for enduring rigour. Similarly, artificial thinking could be of different nature than human thinking.

Correct thinking is one of the main attributes that we identified in the previous subsection when dwelling upon the question what intelligence is.
In his 1950 paper \emph{Computing machinery and intelligence} (\cite{Turing:1950:Intelligence}) Alan Turing started out with the question ``Can machines think''. Being the astute scientist, Turing did not answer the question what thinking exactly is but rather formulated his so-called \emph{Turing test}: can humans tell a conversation with a machine apart from a conversation with a machine? Edsger Wybe Dijkstra (who won the Alan Turing award in 1972) followed the discussion with his famous quote \cite{EWD898}
\begin{quote}
The question whether a machine can think is as relevant as the question if a submarine can swim.
\end{quote}

In this paper we will also not provide a definition of Artificial Intelligence. We recognise it when confronted with it and we shall see various examples. We do find it useful to make a distinction between different types of AI later on in this section. The exact meaning of AI is much in movement. Some beholders only call something AI whenever they are utterly impressed and amazed. However, we should not forget that simple algorithmic tasks like sorting, comparing, and computing are also considered to require a minimum of intelligence. 

Even if there is no consensus about a precise definition of Artificial Intelligence, all scholars do agree that for it to be artificial it needs to be external to humans. Typically it means being performed by a machine. 

\subsection{Should we use Artificial Intelligence?}

Some social scholars point out how Islam and Christianism have dealt differently with the scientific method and critical thinking observing that Christianism has eventually embraced it while Islam has not. In this discussion, the Christian philosopher St. August can be seen as an important exponent to unify the scientific method with Christian dogma. On the other hand, philosophers like Al Ghazali, Ibn Taymiyya and Nizam al Mulk are mentioned as important exponents that gave mysticism more important weight in Islam (\cite{TaymiyyaAgainstGreeks}).

One may wonder if the acceptance and embracement of Artificial Intelligence is of similar nature as the just mentioned embracement of scientific method. If so, the question whether to use it or not is of utmost importance. In this paper it is not really relevant what position to take. We simply observe that AI is being used and there is a strong tendency to use it where possible. As such, we simply mention, that if it used, then it better be used properly. 

Since the field of AI is so much under development and in movement, it may seem natural to not look for an all encompassing overarching legal framework to govern AI, a top-down regulation so to say. As such, it may seem way more natural to start with a bottom-up regulating approach, isolating case-by-case and dealing with them separately to only later look for a more uniform approach.

\section{AI in public administrations}

As mentioned, Artificial Intelligence encompasses a great deal of methods: Machine Learning, Data Querying, Automated Search, Computations, Big Data Processing, Large Language Models, etc. Each method comes with its own particularities, its own benefits and its own shortcomings. It seems difficult to deal with all of these methods in a single treatment. We will distinguish between roughly two types of AI in this paper: Data-driven AI like Machine Learning and Neural Networks and Rule-based AI like in old-style programming.

\subsection{Rule-based AI and Formal Methods}

At the current fast-developing stage of AI, it seems to make sense to consider each AI method apart when applying them in the public administration. Also, the nature of the applications are important. Should AI make decisions? Or should AI merely help to organise information for the human actor? Depending on different purposes, different AI techniques can be more or less suitable. 

An important factor here is whether AI applications in the public administration imply important choices to be made. For example, important decisions will be made on the basis of DNA sampling software. Substantial fines will be issued on the basis of tachograph software. Biometric recognition software may be applied in a way that does not imply important decisions but it may also be used in such a way that it does imply substantial automated decisions. 

In the case that important decisions follow from AI applications it seems indispensable to have some sort of control or knowledge what the AI application has done. We would like to understand the AI decision and be able to explain it to the affected citizens. 
In this sense, various methods will simply disqualify themselves. Control and transparency for Neural Networks seems simply impossible, certainly with the current state of affairs.\footnote{It is actually quite hard to sustain with theoretical results the claim that Neural Networks allow for less control than Rule based AI. One can point at theoretical results to the extent that all important questions about Neural Networks are undecidable \cite{Wiklicky:1994:LoadingProblemUndecidable, UnderfittingUndecidable}. However, this can countered by the Halting problem for Algorithmic AI which is also undecidable. One may then object that learnability may be of a different level of undecidability possibly invoking the continuum hypothesis \cite{BenDavid:2019:LearnabilityUndecidable} but the setting is very artificial and not too convincing. A pragmatic argument is that for Algorithmic AI there is lots of formal methods around \cite{Oregan:2017:GuideFormalMethods} whereas for Neural Networks there are only some tentative starts \cite{Corsi:2021:FVforNueralNetworks, FultonPlatzer:2018:SafeLearning}. It seems however that currently we lack a theoretical framework to evaluate the claim and intuition that rule-based AI allows for more control than data-driven AI. An interesting aside is that Neural Networks seem to adapt to test-phases (\cite{Hutson:2024:TwoFacedAI}) while acting differently outside the test-phases.}

However, deterministic algorithms that follow rule-based
legal reasoning do allow us to get a certain level of understanding. Moreover, we would like to stress that this level of understanding can be stretched to important levels of certainty and transparency if \emph{Formal Methods} are used \cite{Joosten:2023:Dialogues, Oregan:2017:GuideFormalMethods}.
Formal methods are mathematics and logic applied to algorithms and programs to obtain knowledge with mathematical rigour about the behaviour of these programs.

Thus, if control and full transparency is needed in AI applications, Formal Methods are indispensable. Formal Methods are only available however for rule-based AI. What do we mean by rule-based? At the end of the day, algorithms for Neural Networks are also rule based with instructions like: ``if the success rate of the network is in this range then in/decrease the weights of certain nodes in the network''.  When we speak of rule-based AI, we mean those applications of AI where the algorithm performs a task that is based on rules that follow the logical structure of the functional description of the algorithm; Old-style programming as it were.

To sum up this subsection: If we wish to have a substantial amount of control over what AI does, this can only be obtained using formal methods. Formal methods are only available for rule-based AI.


\subsection{Formal ontologies and limitations}\label{section:FormalOntologies}

So far we have argued that if control and full understanding is needed of AI used in the public administration, then we need formal methods and thus should restrict to Rule-based AI. Only the use of formal methods can guarantee full mathematical rigour and certainty. But there are some caveats that one should bear in mind. 

The {\bf first caveat} has to do with formal ontologies. Mathematics and logic deal with ideal objects like numbers, vectors, propositions, etc. When we say that we have full mathematical rigour in the control of the behaviour of the algorithms, this rigour is with respect to the idealised objects like numbers, vectors and the like. One may question whether these objects like number really `exist' in the real world, whatever that would mean, but for sure, their aspects, like cardinality and ordinality, are manifest in our perception of the real world. 

How numbers relate to the real world cannot be controlled with full mathematical rigour. In particular, `understanding algorithms to the full' and formal methods can only apply when there are very well defined formal ontologies. An example of such an ontology is \emph{speed}. Speed can be measured and then can be represented by a number. A less optimal ontology would be \emph{driving} since this does not directly correspond to a physical observable: for example, a driver can be waiting for a traffic light and hence has speed zero yet is driving. Also, the engine activity would not be an optimal modelling choice to signal driving: a driver might be sleeping for half an hour on a very hot day with the air conditioning on and the motor running in stand-by. 

Therefore, if we wish to apply automated reasoning it is imperative to work with a clear set of well-defined ontologies that directly correspond to observables. It are modelling choices and claims that relate the observables to terms in natural language. For example one could choose to let the term \emph{driving} in natural language be flagged by the observable \emph{speed} although, as we have argued, this is not an optimal choice. 

To use formal methods, we thus need to translate legal text through our formal ontologies into a formal language. In the formal language we can reason with precision. This precision is only with respect, however to the \emph{formal semantics}. Formal semantics is the meaning of sentences in the formal language with respect to mathematical idealised objects (like numbers, vectors, etc.). 

Let us give an example. Suppose the formal language is the simple language of propositional logic. So, our language speaks about propositions $P,Q,R,S,\ldots$ that can be combined using connectives like ``and'' (we write $\wedge$), ``if then" (we write $\to$), etc. Propositions can be seen as container places for information. Their formal semantics says\footnote{In classical logic. There are tons of alternative propositional logics.} that the value of a proposition can be either \emph{true} or \emph{false}. Often one writes $1$ for true and $0$ for false in the formal semantics. Now consider a legal stipulation that says
\begin{quote}
If you drive in an industrial area and it rains, your maximum allowed speed is 35 kilometers per hour. \ \ \ \ \ \ \ \ \ \ \ \ \ $(\dag)$
\end{quote}
We can choose our formal ontology as 
\begin{align*}
P & := & \mbox{You drive in an industrial area}\\
Q & := & \mbox{It rains}\\
R & := & \mbox{The driving speed is not more than 35 km/h}\\
\end{align*}
Under this translation we can express the legal stipulation as
\begin{equation}\label{eq:formalLegality}
(P \ \wedge \ Q) \ \to \ R .
\end{equation}
Clearly, we would like to agree that \eqref{eq:formalLegality} is true if and only if the driving behaviour with respect to this clause $(\dag)$ is legal.

We can imagine that $P$ (driving in an industrial area) can be related to the observable of a GPS location. Likewise, $R$ can be related to the values provided by a speeding device. However, $Q$ (it is raining) seems essentially subject to a judgement: would a high humidity with water particles in suspense in the air or very slowly swirling down count as rain? On the formal semantics, there is no ambiguity issue. The value can be either 1 or 0 and which of the two will be provided by the data that we feed to the formal reasoning.  

But regardless of this link of the propositional variables $P, Q$ and $R$ to the real world, on the formal side we can see with full mathematical rigour that all of the following\footnote{The reader is encouraged to translate these so-called \emph{tautologies} back to the legal stipulations in natural language.} are valid: 
\begin{align*}
R \to ((P \ \wedge \ Q) \ \to \ R),\\ 
\neg Q \to ((P \ \wedge \ Q) \ \to \ R),\\
\neg P \to ((P \ \wedge \ Q) \ \to \ R),\\
(\neg P \vee \neg Q)  \to ((P \ \wedge \ Q) \ \to \ R),\\
\mbox{ etc.}\\ 
\end{align*}

This example, should show and clarify that mathematical precision can only be obtained for the relation between formal language and formal semantics. We illustrate this discussion and observation in Figure \ref{figure:formalOntologies} below.

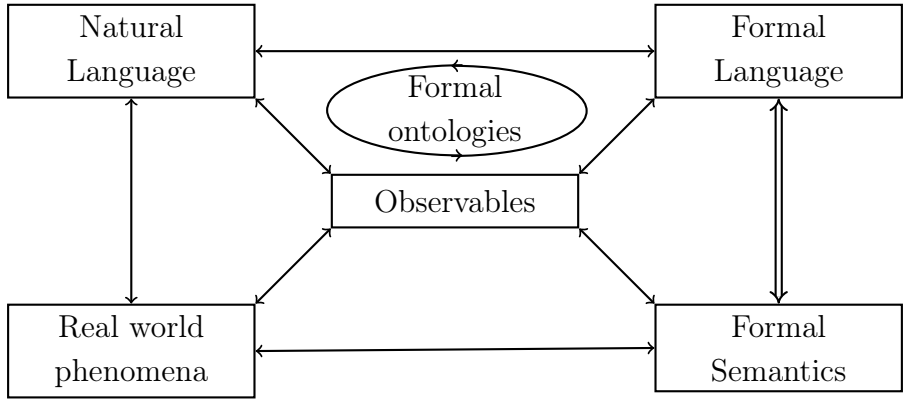
\begin{figure}[H]

 \begin{center}
\tikzset{dbl/.style={double,
                     double equal sign distance,
                     implies-implies,
                     shorten >=0pt,
                     shorten <=0pt}}

\begin{tikzpicture}[
rectanglenode/.style={rectangle, draw=black,thick, minimum size=7mm, text width=3cm, align=center},
]
\node[rectanglenode]      (maintopic)                              {Observables};
\node[rectanglenode]        (formal)       [above right=of maintopic] {Formal Language};
\node[rectanglenode]        (natural)       [above left=of maintopic] {Natural Language};
\node[rectanglenode]        (phenomena)       [below left=of maintopic] {Real world phenomena};
\node[rectanglenode]        (semantics)       [below right=of maintopic] {Formal \\ Semantics};
\node[text width=1.8cm, align=center] (A) at (0,1.2) {Formal ontologies}; 

\draw [->,thick, rotate=-90] (-.6,-0.05) arc [start angle=0, end angle=180, x radius=.6cm, y radius=1.8cm];
\draw [->, thick, rotate=-270] (1.808,-0.1) arc [start angle=0, end angle=180, x radius=.6cm, y radius=1.8cm];

\draw[<->,thick] (formal.south west) -- (maintopic.north east);
\draw[<->,thick] (natural.south east) -- (maintopic.north west);
\draw[<->,thick] (phenomena.north east) -- (maintopic.south west);
\draw[<->,thick] (semantics.north west) -- (maintopic.south east);
\draw[<->,thick] (natural.south) -- (phenomena.north);
\draw[dbl,thick] (formal.south) -- (semantics.north);
\draw[<->,thick] (phenomena.east) -- (semantics.west);
\draw[<->,thick] (natural.east) -- (formal.west);
\end{tikzpicture}

\caption{Formal ontologies should be chosen carefully. Mathematical rigour can only be obtained for the relation between expressions in a \emph{Formal Language} with respect to the \emph{Formal Semantics} of this formal language. This relation is the only one depicted with a double arrow. All other relations are meaningful on the basis of trust and common sense and formal methods in full for them are impossible.}\label{figure:formalOntologies}
 \end{center}
 \end{figure}

We were discussing caveats to the claim that formal methods provide full mathematical rigour. As {\bf second caveat} we mention a position of ours that is much related to the above discussed matter on formal ontologies. We think that a software should not be entrusted with discretional powers. Admittedly, these discretional powers will follow regularities in practice. However, if you catch those regularities in a set of rules you seem to contradict the mere definition\footnote{Collins online dictionary, October 2023.} of discretional power: 
\begin{quote}
Discretionary things are not fixed by rules but are decided on by people in authority, who consider each individual case. 
\end{quote}
When patterns in applying discretional powers are captured by rules and applied in an automated fashion this seems to inevitably give rise to what is known as \emph{fettering}. 

A third {\bf caveat} to the claim that formal methods provide full mathematical rigour has to do with reasoning. We should recall that formal methods rely on mathematical reasoning. In the end it is \emph{us} who reason and humans do make mistakes. This caveat can be mitigated by using so-called \emph{proof assistants} like \emph{Coq}, \emph{HOL}, \emph{Isabelle},  \emph{Lean}, etc. Those proof assistants are small computer programs that check our reasoning. Using proof assistants is very costly and supposes an approximate mitigation of this third caveat since the very proof checking program itself may contain errors, bugs, reasoning or design mistakes, etc. 

There is a minor academic issue here too, namely that the reasoning systems that we use themselves may be flawed. Ever since Gödel we know that in a sense, we cannot fully rule out the possibility that our main reasoning systems are inconsistent and flawed altogether.

A fourth {\bf caveat} to the claim that formal methods provide full mathematical rigour is based on implementations. If formal systems are to be used, this mostly makes sense to do so in an automated fashion. This automation relies on a battery of IT solutions like, processors, interfaces, middleware, etc. 

Notwithstanding the caveats that we mention here, using formal methods seems to be imperative once among others civil rights are at stake.

\section{Logical Lingerings}\label{section:LogicalLingerings}

As we have observed above, there does not seem to be consensus about what Artificial Intelligence exactly is. Sometimes, it seems that those automated tasks that impress the beholder are considered as genuine AI. On the other hand, those tasks that are more standard are then considered just automatisation. 

Developments in AI are changing rapidly and impressive progress is made lately. This sometimes leads to speculations about what AI can do in a (near) future. In this section, we shall see natural limits to AI that follow by mere logic. This should not come as a surprise since all kinds of the well known paradoxes can be cast in the realm of AI. For example:
\begin{quote}
Can AI invent a riddle it cannot solve itself?
\end{quote}
One could understand that inventing a riddle includes providing its solution so that this is an outright paradox. Otherwise, the example provides a limitation since either AI can indeed invent a riddle but then cannot solve it, or it can solve the riddle it invented and then it could not provide a riddle it could not solve.

In the remainder of this section we shall see a classical example of more computational nature and some consequences. So, the purpose of this section is to temper claims like that AI will become omnipotent and omnipresent. The example that we shall see is known as Turing's famous \emph{Halting problem} (\cite{Turing:1936:ComputableNumbers, Kleene:1952:IntroductionMetamathematics, Davis:1958:ComputabilityUnsolvability}). The Halting problem asks whether a program on a certain inputs halts/terminates. We will see that this problem is in general unsolvable: with or without Artificial intelligence; simply impossible.

\subsection{Halting programs and looping programs}

To get a feeling for the halting problem, let us see some easy examples of programs here. 
First, let us consider a program $\Pi$ that can take a natural number (those are the numbers $0, 1, 2, \ldots$) as input and that does the following:
 \begin{itemize}
\item
$\Pi$ reads an input number and stores it at the register $y$;

\item
It checks $\$y$ --the value in the register $y$-- and {\bf if} it is larger than $0$, {\bf then} it will makes the value in the register $y$ one smaller\\
(If $\$ y>0$, then $\$y:= \$y-1$)\\
\ \\
{\bf otherwise} (that is, if $\$ y =0$) it will HALT.
\end{itemize} 
Let us see for example how $\Pi$ will act when we give it the input $3$. Then, a run of $\Pi(3)$ would go as follows:

\begin{itemize}
\item 
In the first step $\Pi$ reads the input and stores it at the register $y$ so that
$\$ y = 3$;

\item 
In the next step, $\Pi$ checks if $\$ y$  is larger than 0. Since $3>0$ it proceeds to replace the current value 3 of the register by $3-1$ so that we get
$\$ y = 2$;

\item 
In the next step, $\Pi$ checks if $\$ y$ is larger than 0. Since $2>0$ it proceeds to replace the current value 2 of the register by $2-1$ so that we get
$\$ y = 1$;

\item 
Repeating we get at the next step that
$\$ y = 0$;

\item
In the next step, it checks if $\$ y$  is larger than 0. Since it is not the case that $0>0$ the program will simply HALT by definition of its code. 
\end{itemize}
We conclude that the program $\Pi$ on input $3$ does indeed give rise to a terminating computation. That is, the computation halts and we write $\Pi(3) \downarrow$ to denote this. It is of course easy to see that this program $\Pi$ simply subtracts 1 from the input until hitting 0 when it will halt. Thus $\Pi$ will halt on any natural number as input. Actually, the $\Pi$ that we have described above is more like an algorithm than a program. A program would be an algorithm implemented in one of the standard programming languages like $C$, Pascal or similar. We will refrain from distinguishing programs and algorithms here.

Let us now see an easy program $\tilde \Pi$ that will not halt on any natural number input $x$. We will write $\tilde \Pi(x)\uparrow$ to denote that the program $\tilde \Pi$ on input $x$ does not halt. We also say that $\tilde \Pi$ \emph{loops} on input $x$.

We define the program $\tilde \Pi$ that does the following:
\begin{itemize}
\item
It reads an input natural number and stores it at the register $y$;

\item
It checks $\$y$ --the value in the register $y$-- and {\bf if}  $\$ y =\$ y$, {\bf then} it will makes the value in the register $y$ one bigger\\
(If $\$ y = \$ y$, then $\$y:= \$y +1$)\\
\ \\
{\bf otherwise} (that is, if $\$ y \neq \$ y$) it will HALT.
\end{itemize} 
Since $\tilde \Pi$ can only halt when some value is not equal to itself, we see that the $\tilde \Pi$ indeed will never halt. A run of $\Pi(0)$ would go as follows:

\begin{itemize}
\item 
$\$ y = 0$;

\item 
$\$ y = 1$;

\item 
$\$ y = 2$;

\item 
$\vdots$

\end{itemize}

So far, we have seen a program $\Pi$ that halts on all natural number inputs and a program $\tilde \Pi$ that loops on all natural number inputs. Moreover, in the two examples that we have seen so far it was clear what happened. Sometimes this is not so clear and we will now exhibit a program $\hat \Pi$ with a less predictable behaviour. 

$\hat \Pi$ does the following:
\begin{itemize}
\item
It reads an input natural number and stores it at the register $y$;

\item
It repeats the following:\\
    If the number is even, divide it by two.\\
    If the number is odd and larger than 1, triple it and add one.
    
\item
And only HALT when we reach one.    
    
\end{itemize} 

Let us now see what happens when we feed a particular input to $\hat \Pi$. Thus, a run of $\Pi(3)$ (starting with $\$ y = 3$) would look as follows:

\begin{itemize}
\item 
$\$ y = 3$;
Since the value is not 1, the program does not halt. Rather it checks whether $\$ y$ is even or odd. Since $3$ is odd, the program proceeds to triple it and adding one to the result ($3\times 3+1 =10$) making this the new value of the register content.

\item 
Thus, at the next step we start with
$\$ y = 10$. Since this is not 1 and since it is even, the program will halve the value.

\item 
$\$ y = 5$;
Since this is not 1, and since it is odd, the program will triple it and add one to it. Thus, in the next step we will start with $3\times 5 +1 =16$.

\item 
$\$ y = 16$; Since 16 not 1, and since it is even, it will be divided by two arriving at:

\item 
$\$ y = 8$;
Since 8 not 1, and since it is even, it will be divided by two arriving at:

\item 
$\$ y = 4$;
Since 4 not 1, and since it is even, it will be divided by two arriving at:

\item 
$\$ y = 2$;
Since 2 not 1, and since it is even, it will be divided by two arriving at:

\item 
$\$ y = 1$; Since now the value equals 1, the program instruction tells the program to HALT.

\end{itemize}

For this particular program $\hat \Pi$ and this particular input $3$ we have seen that $\hat \Pi (3) \downarrow$, that is $\hat \Pi$ on input $3$ eventually halts. It is a mathematical open problem (Collatz conjecture) whether $\hat \Pi$ halts on every natural number input \cite{Collatz}. The program $\hat \Pi$ has been tested for astronomically large input values $x$ and always it has been observed that $\hat \Pi (x)\downarrow$. However, it has not yet been proven that $\hat \Pi$ will halt for all possible inputs.

\subsection{The Undecidability of the Halting problem}

Above we have seen a program $\Pi$ that halted on all inputs, a program $\tilde \Pi$ that looped on all inputs and a program $\hat \Pi$ for which it is currently still unknown if it halts on all inputs or not. Before we will give a proof sketch of the Undecidability of the Halting problem, let us first see why it is important to know whether a program halts on a certain input or not.

Suppose we have a computer program $\Pi$ that computes tax return. So, if the program $\Pi$ is given some revenues file $x$ of a citizen, then $\Pi$ should compute $\Pi(x)$ which is the amount of tax to be returned or to be paid. We would like this program $\Pi$ to halt at a certain stage and give the answer. If we want to fully understand how $\Pi$ works, we would at least like to know that $\Pi$ will halt (and provide an output). The theorem about the unsolvability of the Halting problem states that in general it is not possible to tell whether an arbitrary program $\Pi$ halts on an arbitrary input $x$. Of course, for a particular program $\Pi$ with a particular input $x$ it may be possible to prove that this particular program $\Pi$ halts on this particular input $x$. Hopefully indeed the tax return program will always halt and we can attest to this.

An example of a program that better never halts could be an operating system. So, if $\Pi$ is now the operating system (for example of a smart phone) and $x$ is some begin-configuration (for example, the configuration it had when bought in the shop), then we would prefer that $\Pi(x)$ would never stop and that the operating system keeps on running forever (to keep managing the phone). 

If we wish to have some control on software and what it does, in particular, we should be able to tell whether a program halts on a certain input. The next theorem tells us that this is in its unrestricted form simply impossible. 

\begin{theorem}[Unsolvability of the Halting problem]
There cannot exist a computer program $H$ that solves the halting problem telling whether an arbitrary computer program $\Pi$ will eventually end its computation when it starts on input $x$.
\end{theorem}

Again, we stress that the impossibility of such a program $H$ does not depend on our current ignorance. It is a logical hurdle that can simply not be taken. Advanced AI cannot help with this; not now and not ever in the future. We will now proceed to give a proof sketch of the theorem. 

The proof will go by a so-called \emph{reasoning by contradiction}. Thus, we assume there existed a program $H$ that for any inputs $X$ and $Y$ tells me whether the program $X$ will halt on input $Y$. After some reasoning we arrive at an impossibility so that we conclude that our original assumption (there exists such an $H$) must actually be wrong: there cannot be such a program $H$.

Suppose there existed a program $H$ that given another program $X$ and given an input $Y$ to $X$ tells us whether the program $X$ halts on input $Y$. Without loss of generality we will assume that $H$ will output the value 1 in case the program $X$ halts on input $Y$ and that $H$ will output the value 0 in case the program $X$ loops on input $Y$. The program  $H$ thus takes two arguments and we write $H(X,Y)$ for the outcome of $H$.  In short, we assume that there existed a program $H(X,Y)$ with the property that
\begin{equation}\label{absurdHaltingAlg}
H(X,Y)=
    \begin{cases}
        1 & \text{if } X(Y)\downarrow\\
        0 & \text{if } X(Y)\uparrow .
    \end{cases}
\end{equation}

We emphasise that the program $H$ has two inputs, a program\footnote{The fact that $X$ may not be a program or that $Y$ is not a valid input for $X$ are technicalities that are not essential for the argument.} $X$ and an input $Y$ for the program $X$. For our current generation the fact that a program can also be an input is not really strange since we are used to computers. For example, $H$ could be an operating system, $X$ and app that we downloaded and $Y$ an input for the app $X$. The operating system $H$ would then load $X$ in the working memory, feed it input $Y$ and run how the program $X$ should perform on input $Y$. Thus, the program $H$ has as input another program $X$. For generations before us this was a bit of a mental twist. But again, we are so used that a program in the end is just stored as a binary number. In particular, we are fine with feeding the code of a program $X$ (as binary number) as input to the program $X$ itself; this situation would be denoted by $X(X)$. It is exactly this trick that we are going to play now to get some self reference going on.

Thus, using the alleged halting program $H$ it is not hard to see that we can tweak $H$ to get a program $\tilde H(X)$ with just one argument $X$ with the property 
\[
\tilde H(X)=
    \begin{cases}
        0 & \text{if } X(X)\uparrow\\
        \uparrow & \text{if } X(X)\downarrow .
    \end{cases}
\]

How would we make the program $\tilde H$? We want $\tilde H(X) =0$ in case $X(X) \uparrow$. How can we decide whether $X(X)\uparrow$? Of course, we use the program $H$ to this end and simply run $H(X,X)$ and we will program $\tilde H$ so that it outputs a 0 in case that $H(X,X) = 0$.

However, in case $H(X,X) =1$ we are going to do something special: we make $\tilde H (X)$ enter in a silly loop. For example, through executing the program $\tilde \Pi$ from the previous subsection. The thus described algorithm gives us the new program $\tilde H$. 

So, $\tilde H$ is a determinate program and we can feed it to itself as input, that is we can evaluate $\tilde H(\tilde H)$. But now, we get a contradiction since in case  $\tilde H(\tilde H) \downarrow$ this can only be in case $\tilde H(\tilde H) =0$ and then, by definition of $\tilde H$ this is exactly the case when $\tilde H(\tilde H) \uparrow$. And vice-versa, if $\tilde H(\tilde H) \uparrow$, then by the definition of $\tilde H$, this can only be when $\tilde H(\tilde H) \downarrow$. This is clearly a contradiction since we have concluded
\[
\tilde H(\tilde H) \downarrow  \ \Longleftrightarrow  \ \tilde H(\tilde H) \uparrow.
\]
Thus, our original assumption that there existed such a program $H$ with the properties as stated in \eqref{absurdHaltingAlg} lead to a contradiction and we conclude that there exists no such algorithm. This completes the proof sketch of the undecidability of the Halting problem. 

The Halting problem and related problems turn up in various real life problems. A direct corollary of it is that there cannot be an infallible virus-scanner; with or without AI. Likewise, the Halting problem yields that unrestricted program correctness verification (what we so much want for software that affects citizens) is impossible. Again, all this impossibilities hold with or without AI. Simply, these problems are not computable.

\section{Computational Complications}\label{section:ComputationalComplications}

In the previous section we have seen limits on computational tasks. These limits are naturally imposed by mere logic itself. We saw that certain problems, like the Halting problem, are what is called uncomputable: there cannot exist a computer program that solves the problem in its full generality. AI cannot help at all here.

We thus have computable and uncomputable problems. Clearly, the uncomputable problems pose a limit on AI and applications. In this section we will see that even in the realm of computable problems things can be bad. Some problems are computable in the sense that we can in principle solve them: we know a methodology to solve the problem. However, this methodology may essentially need more computational resources than available in the entire visible universe. 

One may think that those problems are artificial and do not pop up in real life. This is unfortunately wrong. Computable but unfeasible problems are actually abundant in real life. Let us imagine there is the following law: 

\begin{quote}
Upon matrimonial divorce, each item in the common patrimony should be assigned a € amount and then the patrimony should be split between the partners in such a way that causes the minimal € difference in the division of goods between the partners.
\end{quote}

The law looks fairly decent and one can imagine such a law being in force in some society. Moreover, we can imagine that it will be a computer that performs the task of finding an optimal distribution of the goods between the divorcing partners. Then, as legal supervisors we would like to be sure that the computer has been fair in its decision. We now stand for an unfeasible taks: there are typically too many  possible distributions. If there are $n$ items in the common patrimony, then there are $2^n$ many ways to distribute the goods between the divorcing partners. Recall that $2^n$ stands for $\overbrace{2\times \ldots \times 2}^{\mbox{$n$ times}}$ and that this grows pretty fast: with $2^2 =2\times 2 =4$, with $2^3 =2\times 2\times 2 =8$, with $2^4 =2\times 2\times 2 \times 2 =16$, etc. 

If the partners have 1000 items in the common patrimony, there are $2^{1000}$ many possible distributions of the items between the divorcing partners. The legal authority/supervisor would have perform $2^{1000}$ many checks. This is impossibly huge. To give an impression, let us write this out.
\[
2^{1000} = 
\]
10715086071862673209484250490600018105614048117055336
07443750388370351051124936
12249319837881569585812759467291755314682518714528569
23140435984577574698574803
93456777482423098542107460506237114187795418215304647
49835819412673987675591655
43946077062914571196477686542167660429831652624386837205668069376.

Okey, one may say that's a big number allright. However, we now have supercomputers, we have AI and we have all kinds of fancy computing devices and support. But all of this will not help. To given an impression: scientists estimate that the number of elementary particles in the visible universe does not exceed $10^{100}$. So there is simply not enough matter in the entire universe to write down all the possible options to distribute the items. So memory space just runs out. But even if we had some smart way of not having to write everything down, there is simply not enough time to check all distributions. To give an idea, there are a bit less than 32 million seconds in a year. With the estimated age of the universe of around 15 bilion years,  there have elapsed less than $10^{18}$ seconds since the big bang. You see, we are not even getting close. 

The astute reader may think that there is a clever way to find the optimal distribution that will not require just checking all $2^{1000}$ possible distributions. For example, we do not need to consider the distributions where one of the partners gets everything and the other gets nothing. However, such simple considerations do mostly not substantially reduce the complexity of the problem. In particular, theoretical computer science tells us that there are tons of problems like the one above where one can mathematically prove that no smarter solution exists than roughly checking all exponentially many possibilities. In other situations this is only conjectured with the most famous example being the problem whether $\mathbf P = \mathbf{NP}$.

Problems like this show up in many places. Theoretical computer science can prove (\cite{Wegener:2005:ComplexityTheory, Papadimitriou:1994:ComputationalComplexity}) that various problems are not solvable in realistic time: with or without AI. Whereas the uncomputability from the previous section formed a clear boundary to AI, in this section we have seen that various problems are solvable in principle but in practice they are not feasible: they can not be computed using all available computational resources. 

\section{Mathematical Mischiefs}

In the previous sections we have seen problems of AI that were of a fundamental nature: certain computational problems are not computable at all, or computable but requiring an astronomically large amount of computational resources. Let us call this class of problems \emph{fundamental AI problems} for the moment. We shall see that fundamental problems do not evidently manifest themself so often. Then we shall consider a class of problems/issues in AI that we so far not yet considered: AI problems/issues that are due to underspecification, ambiguity, or unexpected dynamics in the natural language law.

\subsection{Fundamental AI problems rarely show up}

Fundamental AI problems do not manifest themselves often in society. This is for good reason. If an AI application happens to need to solve a fundamental AI problem, the users of the program will just find out that the program gets frequently stuck and engineers will find some other approach that avoids the fundamental AI problem (possibly by simply ignoring the technical specs if needed). 

It can happen though that a fundamental AI problem only manifests itself in very rare occasions. In such a situation, the fundamental AI problem may go by unnoticed for a long time. In particular, the AI problem may not manifest itself during the development and testing stage of the software. One can then only hope that when the fundamental AI problem does manifest itself later, this is not in a critical situation that can lead to much damage or even casualties. Another argument why formal methods are important: to warrant good behaviour of the program at the level of logic and in particular to warrant that the program is free of fundamental AI problems. However, in most AI systems, there will be no casualties nor much damage in case of malfunctioning. So in most of the cases, if a fundamental AI problem does manifest itself, the user of the AI application will notice that the "system gets stuck", get annoyed and probably reset. Often, the fundamental AI problem will disappear with a different set of parameters after the reset and the fundamental AI problem has gone by unnoticed.

Summing up, either a fundamental AI problem often manifests itself in an application so that the engineers will remove it (possibly ignoring the program specs) or the fundamental AI problem in an application only manifests itself once a blue moon in which case the user will try to find a work-around possibly by simply resetting. In both cases, a fundamental AI problem will not be consciously noticed and recognised as such by the user.

The kind of problems that we most often see in automated decision making is of a different nature. In this paper, we collect those problems by the name of \emph{mathematical mischiefs}. These are situations where the law or specification implies bad mathematical behaviour. Often this is behaviour that was not even realised to be present in the law/specification by the programmers or those who draft laws.

\subsection{Mathematical mischiefs}\label{section:mathematicalMischieves}

It is common practice that law is written in natural language even if this law is meant to be enforced in an automated fashion. The process \emph{From Law to Code} is much discussed in the literature and in general it still remains unclear what optimal practices are. Some projects opt for semi-natural-languages like the Catala project \cite{MerigouxEtAl:2021:Catala, HuttnerMerigoux:2022:CatalaFuture} but natural language is still the most commonly accepted mode.

Especially if law is written in natural language and if that law deals with numerical stipulations, then this law is typically very prone to imply possibly rather wild mathematical or otherwise undesired behaviour. In the next section we will illustrate this by giving various examples of mathematical problems in law. Many other examples are around in the literature \cite{merigoux:2023:Rules, Errezil:2019:Homologation} but here we shall mainly take examples from\footnote{Some examples are not in \cite{mullerJoosten:2023:modelcheckingPreprint} but only in an unabridged online version of it \url{http://www.joostjjoosten.nl//papers/2023ModelCheckingInFoundationsComputableLaws.pdf}.} European traffic regulations as described in \cite{mullerJoosten:2023:modelcheckingPreprint}. We shall see that some of our examples could also have remained unnoticed in case the law were written in (pseudo)code. Thus, switching to pseudo-code will not be a medicine for all mathematical mischiefs.

\section{Multiple Mathematical Mischiefs}\label{section:MultipleMathMischief}

In this section we shall exhibit various peculiarities of one single regulation: Regulation (EC) no 561/2006 of the European Parliament and of the Council of 15 march 2006 on the harmonisation of certain social legislation relating to road transport.

The main purpose of the traffic regulation is to warrant road safety. The regulation aims to contribute to safety on public roads by imposing enough rest-periods between driving for, among others, truck drivers. The activities of the drivers are being recorded by a so-called tachograph. For the sake simplicity and exposition we will consider here only a small selection of articles from the regulation and omit others. Thus, we will consider a single driver whose time dedication will be labeled either as \emph{driving}, \emph{resting} or \emph{other work} which will also be archived on a so-called \emph{driver card} of the truck driver. Legal issues and problems are known to occur often in the realm of temporal (computable) laws \cite{MonatMerigoux:2024:DatesAmbiguitiesLaw}.

The peculiarities that we point out in this paper all fall mainly in the category of \emph{Mathematical Mischiefs} originating from undesired interactions between sorts that are of fundamentally different nature. For example formal language versus natural language, formal ontology versus informal ontology, legal language versus natural language, etc. We will try to classify the mischiefs by using different subsections.

\subsection{Underspecification}\label{Mischiefs:Underspecification}

A classical source of trouble with computable laws can be called \emph{underspecification}. A legal article $A$ will stipulate certain requirements to comply with the law. Then there will often be some situation $S$ where it is unclear weather $S$ is legal according to the article $A$ or not. Sometimes, the underspecification in $A$ is evident, sometimes it is more subtle. We will begin with a simple example of certain subtlety. To this end we first give the relevant article which is Article 6, Point 1. 

\begin{quote}\sf
Article 6.1: The daily driving time shall not exceed nine hours. However, the daily driving time may be extended to at most 10 hours not more than twice during the week.
\end{quote}

At first sight, the article looks unproblematic. To spot the underspecification is a bit subtle. To see the problem of Article 6.1, we first need to comment on some of the words in the article.  

First, there is word \emph{week}. The regulation in its Article 4 (i) is very explicit in what it understands by a week: 

\begin{quote}\sf
Article 4 (i): \ \ ‘a week’ means the period of time between 00.00 on Monday and 24.00 on Sunday. 
\end{quote}
If we really want to be pedantic: Article 4 (i) of the regulation does not explicitly state that the Sunday actually refers to the first Sunday following the Monday under consideration. Oh well, that's not too bad and one can defend that a bit of common knowledge may be assumed. But still, common knowledge is not too stable and the programmer will need to read between the lines if (s)he is to write a program that implements the law for tachograph readers (that clearly will be executed on a computer).

Another pedantic subtlety lies in possibility of so-called \emph{leap seconds}. Leap seconds (\cite{UTCStandard,LeapSecondList}) are seconds that may be added to a day to compensate for the (positive or negative) acceleration of the rotation of the earth. Let us not go into details here but every half year the International Earth Rotation and Reference Systems Service may decide to add or subtract a leap second to our calendar. Thus, a calendar minute can be either 59 seconds (in case of a negative leap second), or 60 seconds (for a regular minute) or 61 seconds (in case of a positive leap second is added to this particular minute). By convention there can be at most two leap seconds per year to be included at the end of the last day of the month, preferably in June or December.

It may thus happen that Sunday either ends at 23:59 or at 24:01. In the first case, what should the programmer do? Due to the negative leap second, Sunday will end at 23:59 and the moment 24:00 on Sunday does not exist? If the programmer is to follow the letter of the law, the week will then end on the next Sunday 24:00? We think that according to the spirit of the law, it is clear how to act. We should understand that Article 4 (i) really means to say: ``A week is the period of time comprised between the start of a Monday and the end of the next Sunday after it.'' Innocent, but indeed the programmer should make a legal decision here. Again, this is a rather innocent example, but it will get worse. To see how the ambiguity present in Article 6.1 gets worse we will have to tell what the regulation understands by \emph{daily driving time}. In Article 4 (k) we learn that 
\begin{quote}\sf
Article 4 (k): \ \ ‘daily driving time’ means the total accumulated driving time between the end of one daily rest period and the beginning of the following daily rest period or between a daily rest period and a weekly rest period;
\end{quote}
Let us continue in the pedantic mode and observe some minor unimportant issues. First we observe that \emph{daily} does not refer to what most people understand by day. If we recall the discussion from Subsection \ref{WhatIsAI:subsectionThenArtificial} on adjectives, we could say that in the phrase ``daily driving time'' we work with a privative rather than a subsective modifier. As we will also observe in Subsection \ref{Mischiefs:OntologiesAndNames} the use of privative modifiers in communication between legal and IT scholars seems to be begging for problems. 

But, there is another underspecification here: What about driving time between two weekly rest periods? Imagine that a driver holds a weekly rest period and then only works one day before taking another weekly rest period. This seems to be a non-standard situation and a situation that the employer would desire not to occur. Nonetheless, the situation can perfectly take place and, according to the letter of the regulation, it is legal\footnote{The nature of our example is shaped to comply with the other requirements imposed by Article 7 and Article 8.2. Here Article 7 says ``After a driving period of four and a half hours a driver shall take an uninterrupted break of not less than 45 minutes, unless he takes a rest period.'' and Article 8.2 says ``Within each period of 24 hours after the end of the previous daily rest period or weekly rest period a driver shall have taken a new daily rest period.''. Moreover, Article 8.3 says ``A daily rest period may be extended to make a regular weekly rest period or a reduced weekly rest period''. It is a bit unclear if you extend a daily rest period to become a weekly rest-period if than that strip of time is then simultaneously a daily and weekly rest period. To have our example work we understood that if a daily rest period is extended to become a weekly rest period, then it ceases to be a daily rest period and becomes only a weekly rest period.} to have a weekly rest be followed by four and a half hours of driving, then 45 minutes rest, then four and a half hours of driving, then 45 minutes rest, four and a half hours of driving, to be followed by a next weekly rest period. 

At first sight, since we squeezed 13.5 hours of driving in a short period, it seems that we have a violation with Article 6.1 as quoted above.
However, since our driving activity is immediately delimited by two weekly rest periods, the driving activity is strictly speaking not defining a daily driving time. Clearly this is not according to the \emph{spirit of the law}. We will give some more examples of this in Subsection \ref{Mischiefs:FarmalisationVsSpirit}. 

Before we go on to expose the most striking under-determinacy  of Article 6.1 let us mention yet another pedantic issue of Article 6.1 in combination with Article 4(k). 
There is a degenerate boundary case that is problematic to Article 4 (k). Namely when a driver is new to the office. When a driver is new to the office and will start driving, the corresponding driver card will not have any activities recorded and in particular will not have a (daily) rest period yet so there cannot be any daily driving time either. Of course, there is an easy and natural way to deal with this slightly academic anomaly. But again, this is an example of a (straightforward in this case) decision left to the programmer/modeller.

By now it should be clear that a daily driving activity can fall in between two days. Now let us consider the situation that a driver starts driving on a Sunday evening, ending on a Monday early morning. So far so good. However, if the accumulated daily driving time in that period added up to 10 hours we see yet another underspecification. According to the regulation it is allowed to have a driving period of 10 hours but not more than twice per week. Since week is defined as starting on Monday 00:00 and ending on Sunday 24:00, Article 6.1 of the regulation is not clear as to which week the extended daily driving time should be attributed: to the week that contained the Sunday where the driving started, or to the week containing the Monday where the driving ended. Is the driver free to choose? Again, the regulation is underspecified here. In the disambiguation presented in \cite{mullerJoosten:2023:modelcheckingPreprint} the extended daily rest period is assigned always to the week that starts on the Monday where the driving ended. Various different tachograph readers seem to make different choices and hence they implement different laws. Some tachograph softwares have an option to fix your choices or to choose the distribution as to minimise the fine. Clearly, all this uncertainty could have been avoided by having unambiguous articles in the regulation. Without the use of formal methods, it is very likely that underspecification still is present somewhere possibly at a hidden or unexpected place.
%
%
%
%
%


\subsection{Inconsistencies}

Legal regulations can contain simple inconsistencies. This can then lead to a situation where there is no legal behaviour at all. We will also speak of inconsistency when there is no reasonable legal behaviour at all. A famous example comes from a 20th century train regulation\footnote{We do not have exact reference to the regulation and it may as well be an apocryphal origin. The example serves a mere illustration.} from Kansas:

\begin{quote}\sf
When two trains approach each other at a crossing, both shall come to a full
stop and neither shall start up again until the other has gone.
\end{quote}

Once two trains will meet at a crossing, it is clear that the trains enter in a deadlock and must stay there forever if they wish to obey the law. Staying put forever is clearly not reasonable. The example from the European traffic regulation that we point out her comes from Article 7.


\begin{quote}\sf
Article 7 (1st part): After a driving period of four and a half hours a driver shall take an uninterrupted break of not less than 45 minutes, unless he takes a rest period.

Article 7 (2nd part): This break may be replaced by a break of at least 15 minutes followed by a break of at least 30 minutes each distributed over the period in such a way as to comply with the provisions of the first paragraph.
\end{quote}

We observe that Article 7.2 strictly speaking is inconsistent in the following sense. The second part of Art. 7.2 describes a situation which is in conflict with the first paragraph but allowed by way of exception. So far so good, but then it says "in such a way as to comply with the provisions of the first paragraph" which we observed is impossible. This is an innocuous inconsistency because everyone will simply tacitly understand that this last phrase 
-- "in such a way as to comply with the provisions of the first paragraph" --should simply be ignored. However, it is a decision that needs to be made to consistently interpret the law and in a sense, it is a free choice up to the programmer or modeller. More subtle examples of the modeller taking essential interpretational decisions are dealt with in \cite{DEALMEIDABORGES2020104636, Errezil:2019:Homologation}.

%
%

\subsection{Mismatch spirit versus formalisation of the law}\label{Mischiefs:FarmalisationVsSpirit}

Often, legal regulations allow situations that are not according to the \emph{spirit of the law} or the purpose of the law. The purpose of our leading example, the European traffic Regulation (EC) no 561/2006 is to make sure that drivers take regular rest periods and do not spend too much time in a row in their truck. Spending too much time in a truck will make the driver tired and sleepy and less alert thereby increasing the chance of traffic accidents. We shall now exhibit a a situation that is legal but where the driver just spends so much time in the truck that we doubt that it is intended to be legal. We think that the authors of this law simply never considered this possibility.

To exhibit our example, we continue with Article 7 which stipulates that driving periods should have rests between them. The question is, \emph{what is exactly a driving period}? Let us consider some different driving patterns.

{\bf Driving Pattern 1} A driver drives for one hour, then stands put for a minute and then drives another hour. 

Should we consider Driving pattern 1 as two distinct driving periods of one hour separated by one minute of rest? Or should we  consider Driving pattern 1 as a single driving period with a duration of two hours and one minute? As humans, we feel that some context is missing: did the driver stand put while being in the truck, for example, while being caught in a traffic jam? If the driver had indeed been caught in a traffic jam, the minute of rest is not really rest and should be counted as work and not as rest. We should remember that this is a case distinction that a tachograph cannot make. The tachograph just measures motion and cannot do much more than that. On the other hand, if the driver really went out the truck to go to the restroom or do some stretching, then we are in a different situation. In this case, we could consider Driving Pattern 1 to consist of two driving periods of one hour, separated by one minute of rest. Again, our human condition speaks up: can one really speak of rest if the duration of the rest period lasted only for a single minute? It seems that the regulators also had this reflection and in another European regulation ((EU) 2016/799) it is stipulated (see o.a. \cite{DEALMEIDABORGES2020104636, FernandezDuqueEtAl:2019:DriveOrNotRegulations50Etc} for details) that any minute of rest that is immediately preceded and immediately succeeded by driving should be regarded driving. Hence, Driving Pattern 1 should indeed be considered as a driving period of 121 minutes. When does a intercalated rest really constitute rest then? According to the law, two minutes is already good. So let us consider the next driving pattern:

{\bf Driving Pattern 2} A driver drives for one hour, then stands put for two minutes and then drives another hour. 

As mentioned, the two minutes of non-movement should not be considered driving but as rest instead. But then, how should we consider Driving Pattern 2? The rest is not long enough according to the above-mentioned Article 7 to really have two different driving periods. It thus seems that we should conceive Driving Pattern 2 as a single driving period with a duration of 120 minutes. This reading is also reinforced by:

\begin{quote}\sf
Article 4 (j) : \ \ ‘driving time’ means the duration of driving activity recorded:
\begin{itemize}
\item[-] automatically or semi-automatically by the record- ing equipment as defined in Annex I and Annex IB of Regulation (EEC) No 3821/85, or
\item[-] manually as required by Article 16(2) of Regulation (EEC) No 3821/85;
\end{itemize}
\end{quote}

Now that we seem to understand how to recognise driving periods and their durations, we consider the next:

{\bf Driving Pattern 3} A driver drives for one minute, then stands put for two minutes and then drives another minute. 

In the light of what we have seen before, Driving Pattern 3 defines a single driving period of a total duration of two minutes. So far, so good, but now we can now take Driving Pattern 3 to the extreme and repeat it very often. Admittedly, it gives rise to an artificial/academic driving pattern but nonetheless it is a pattern that can occur.

{\bf Driving Pattern 4} A driver repeats Driving Pattern 3 for 135 times straight in a row.

Each Driving Pattern 3 adds two minutes to the total duration inside Driving Pattern 4. Thus, Driving Pattern 4 defines a driving period of $135  \times 2 \mbox{ minutes }= 270 \mbox{ minutes } = 4.5 \mbox{ hours}$. So, as far as the regulation is concerned, and in particular Article 7 of the regulation, Driving Pattern 4 is completely legal. However, we doubt that it should indeed be legal, given that the total duration of Driving Pattern 4 is nine hours. The legality seems to violate the spirit of the law since engaging for nine hours straight in the execution of Driving Pattern 4 would most certainly create an exhausted driver prone to engage in a traffic accident.



\subsection{Bad formal ontologies and names}\label{Mischiefs:OntologiesAndNames}

The previous subsection observed that \emph{driving} was never really defined in the regulation. This is particularly doubtful for a regulation that has to regulate exactly that: allowed driving periods and required resting periods. But a moment of thought prompts that actually driving is a complicated ontology indeed. Especially in the case of computable law. Recall that in computable law, the regulations should be evaluated against the data files in an automated fashion. Typically, the data collection also proceeds in an automated fashion as in the case of the tachographs. 

The undefinedness of driving illustrates a more general point. Ideally, legal ontologies in Computable Law should be directly related to formal ontologies in the sense of Subsection \ref{section:FormalOntologies}. Clearly, driving is not an physical observable. There is human judgement involved to decide whether some stretch of time should be considered driving or not. Thus, in the case of Computable Law one should strive for ontologies that are physical observables like speed, weight, etc. If some ontology is not a physical observable, the regulation should tell how the ontology is interpreted in terms of physical observables. As an example, we have seen the poorly designed equation (with some minor amendments) of driving time to moments with positive speed.

We would like to conclude with a comment on the names that regulations in Computable Law use. Of course, some names are to be used. Regulations are still primarily written in natural language (see \cite{Guintchev:2024:LanguageVsPrinciples} for a discussion). However, often it is not clear if a word in a regulation denotes some technical meaning or the common sense meaning. If someone says that she is going for a week on holidays, nobody will think ``ah, you are going from Monday 00:00 through Sunday 24:00 on holidays''. No, rather one will think of a holiday with a duration of seven days. 

Of course, it is fine to have a technical definition of a week in a regulation. However, one should be very clear about this. Regulation 561/2006 is not too clear about it though. It does not help that the regulation also uses the informal meaning of week when using the modifier ``weekly rest period": this is also a technical definition but in most of the cases it will simply coincide with our informal usage of week: a period of seven days.  

One can imagine guidelines of good practice for filing computable laws. For example, if a technical meaning is given to a word used in common language, the usage of the technical term should be flagged somehow. For example by annotating the word where, e.g., $\overline{\mbox{week}}$ would denote to the technical term and week to the common-sense term. An alternative could be to invent new words like `formalweek'. The latter approach has been taken in \cite{AlmeidaBorges:2024:TimeLibrary}. Another good practice would be to avoid the usage of privative modifiers where, for example, `weekly rest period' defines something that has little to do with the meaning of week. There is even a rather amusing article in the regulation that clearly illustrates this.

\begin{quote}\sf
Article 8 (9): \ \ A weekly rest period that falls in two weeks may be counted in either week, but not in both.
\end{quote}
%
%


%
%

\subsection{Hidden mathematical dynamics}\label{section:hiddenStructure}

From mathematics we know that simple rules can give rise to very complicated structures. And from logic in a sense we know that obtaining complicated structures cannot be avoided. Probably the most famous example of extremely complex behaviour arising from a very simple rule is the Mandelbrot set. 
\begin{figure}[H]
 \begin{center}
  \includegraphics[scale=.41]{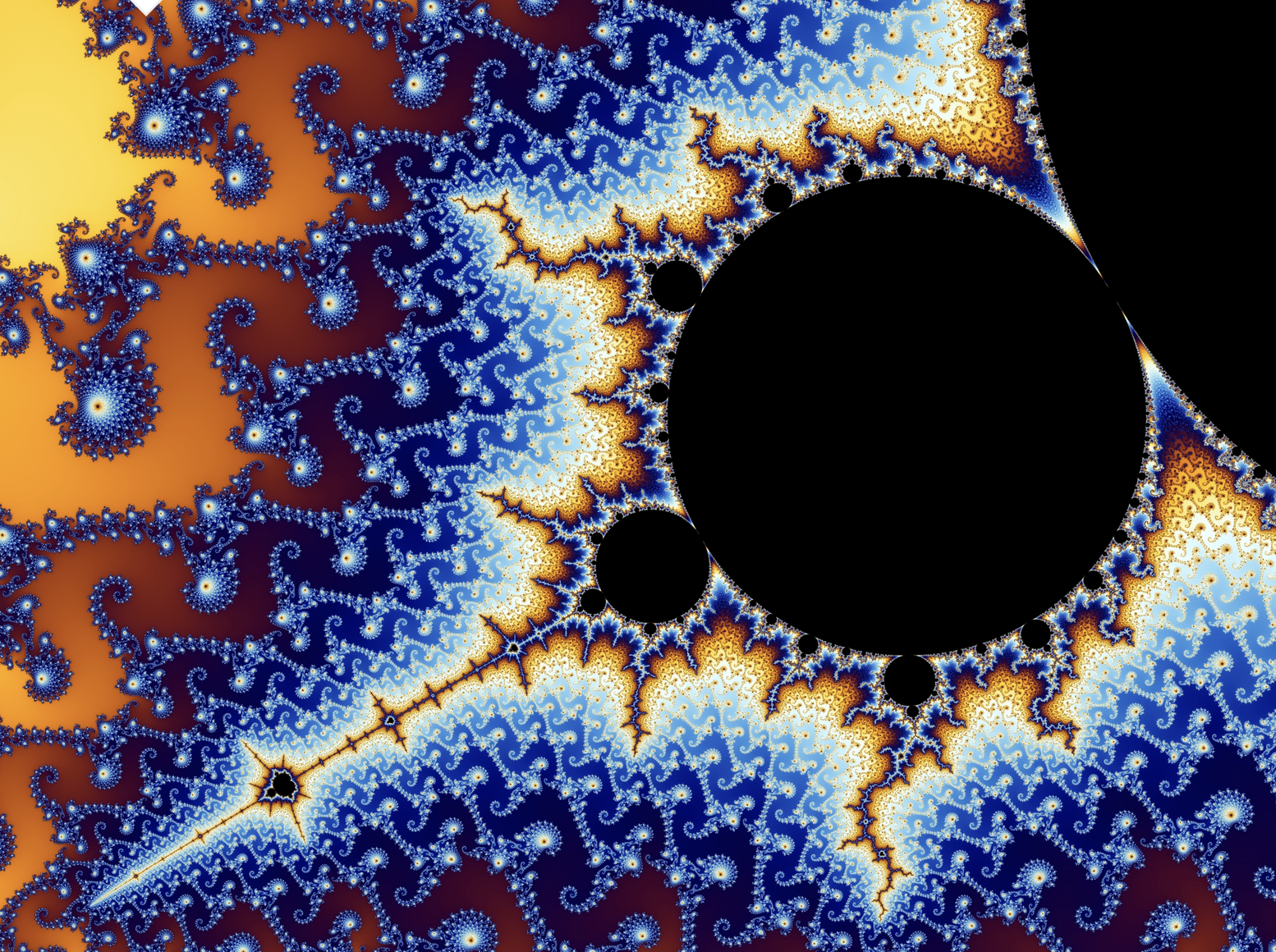}
\caption{The Mandelbrot set}\label{figure:secondToMinuteLabeling}
 \end{center}
 \end{figure}
In a sense, the Mandelbrot set is generated by the following simple rule: Take a number, multiply it by itself and add a constant to it and repeat this. In mathematical notation, the Mandelbrot set arises by iterating $z\mapsto z^2 + c$, a very simple rule indeed. 

Computational Law imposes rules typically on numerical data. These rules then can give rise to unexpected mathematical behaviour. Even though not so complex as the above Mandelbrot set we will now mention two particular examples of complex behaviour that we uncovered from legal rules in traffic regulation. Probably, if the authors of the regulation would have been aware of the resulting complex behaviour, they would have rather chosen a different  regulation.

{\bf Non locality} As first example, we mention a result of the dynamics of Regulation 561/2006 which is peculiar: to test legality of a driver in a particular week, we may need to look what the driver has done five weeks ago, or twelve weeks ago, or any arbitrary number of weeks away from where we are now. We cannot imagine that this is a desirable property of legal regulation. We will not go into the technical details here and rather refer to \cite{DEALMEIDABORGES2020104636}. The non-locality dynamics is ignited by the following articles:

\begin{quote}\sf
Article 4(h): ‘regular weekly rest period’ means any period of rest of at least 45 hours.\\\\
Article 8.6 (1st part): In any two consecutive weeks a driver shall take at least
\begin{itemize}
\item[-]
two regular weekly rest periods, or one regular weekly rest period and one reduced weekly rest period of at least 24 hours. 
\item[-]
 one regular weekly rest period and one reduced weekly rest period of at least 24 hours. However, the reduction shall be compensated by an equivalent period of rest taken en bloc before the end of the third week following the week in question.
\end{itemize}
\end{quote}
The main idea now is that to compensate in a week, one can do so in two weeks later, which by compensating, will need compensation itself and so can carry on arbitrarily far away from the week under consideration. Again, we refer to \cite{DEALMEIDABORGES2020104636} for details, but the idea of the dynamics is nicely captured in Figure 3 below.
\begin{figure}[H]
 \begin{center}
  \includegraphics[scale=.7]{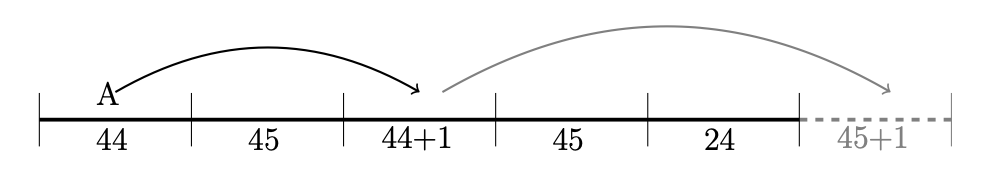}
\caption{Compensating reduced weekly rest periods}\label{figure:WeeklyRest}
 \end{center}
 \end{figure}

In the current paper in Section \ref{Mischiefs:Underspecification} we observe that non-locality also occurs due to Article 6.1.

{\bf Shift Invariance}. As a second example of hidden mathematical dynamics we mention a peculiarity from Regulation (EU) 2016/799 \cite{Regulation799}: tachographs may record differently if you start your minute shifted by some seconds. This seems to be an innocuous feature since one may think that it is perfectly well defined when each minute starts, for example, the first minute of the day starts at\footnote{We use the standard time-stamp format hh:mm:ss where hh denotes the hours, mm the minutes, and ss the seconds of a particular moment in time. Whereas the regulations are very explicit (see Article 4 (i) mentioned above) on the formal definition of what a week is, they do not explicitly state how minutes are defined.} 00:00:00 and ends at 00:59. So where is the problem? 

It may seem a problem that truck drivers  typically will move in between different time zones. For example, crossing the country border from Portugal to Spain will result in an instantaneous leap forward in time of one hour. The European road transport regulations have provisioned for this by requiring the use of a global timestamp that ignores local times: one is to use UTC\footnote{UTC stands for Coordinated Universal Time \cite{UTCStandard} and is a standard that is used worldwide.} timestamps in the recordings of tachographs. 

Now here is the curious issue: the UTC timestamps are defined to include leap-seconds. However, it has been observed and confirmed that at the moment of writing this paper, all tachograph recorders measure in so-called UNIX time. UNIX time ignores leap seconds and at the time of writing this paper there is 27 seconds between UNIX timestamps and UTC timestamps. 

Based on this shift, there exists now a driving pattern $d$  of 5 hours so that $d$ being interpreted as rest or driving will depend on using UNIX time or UTC time. In particular:
\begin{itemize}
\item
If one would use a tachograph based on UNIX time, then $d$ is interpreted as 5 hours rest;

\item
If one would use a tachograph based on UTC time, then $d$ is interpreted as 5 hours of driving.
\end{itemize}

In particular, the driving pattern $d$ will be legal according to UNIX time and illegal\footnote{With respect to Article 7 mentioned above. The driving pattern $d$ is of course rather artificial so as to yield the result. However, real differences of recorded driving time based on UTC versus UNIX are found to be around 
8$\%$, see \cite{AlmeidaBorges:2022:PosterAchtProcent}.} according to UTC time. This is clearly a highly undesirable situation. We will to \cite{DEALMEIDABORGES2020104636} for details. However, the main idea how the strange driving pattern $d$ can be defined should be clear from the relevant regulations and Figure \ref{figure:secondToMinuteLabeling} below. 

Regulation (EU) 2016/799 lays down the requirements for the design and use of tachographs and in particular stipulate that tachograph data recording should happen at the level of the second.
This data is used to determine whether drivers have complied with Regulation (EC) 561/2006 \cite{Reg561}, which is written assuming a minute-by-minute temporal resolution.
In particular, Items (51) and (52) of Section 3.4 of Regulation (EU) 2016/799 regulate how the second-by-second data recorded by tachographs is to be translated into a minute-by-minute format.
  They read as follows:

\begin{quote}\sf
(51): \ \  Given a calendar minute, if DRIVING is registered as the activity of both the immediately preceding and the immediately succeeding minute, the whole minute shall be regarded as DRIVING.

(52): \ \  Given a calendar minute that is not regarded as DRIVING according to requirement 051, the whole minute shall be regarded to be of the same type of activity as the longest continuous activity within the minute (or the latest of the equally long activities).
\end{quote}

In particular, the reference to \emph{latest} second above in Item (52) is not shift-invariant. Based on this one can define a driving pattern $d$ as depicted below so that the driving pattern is interpreted differently depending on using UNIX or UTC time.

\begin{figure}[H]
 \begin{center}
  \includegraphics[scale=.6]{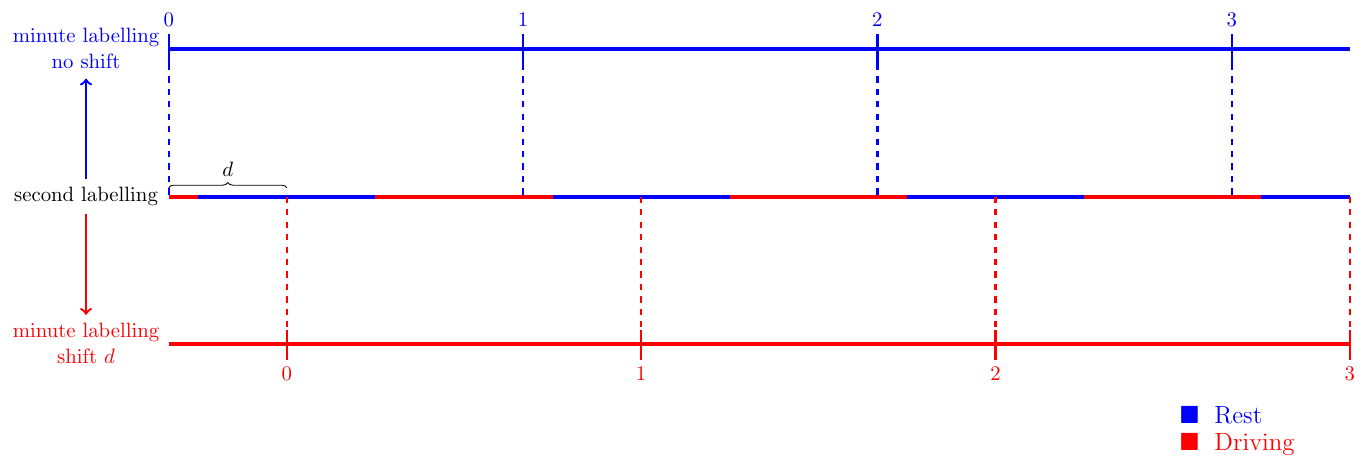}
\caption{In the middle line we have the tachograph recording second by second. The upper line is a minute-to-minute interpretation in case UNIX time would be used. The lower line is a minute-to-minute interpretation in case UTC time would be used.}\label{figure:secondToMinuteLabeling}
 \end{center}
 \end{figure}

\subsection{Shades of Computational Complications and Logical Lingerings}

Article 8.6 above in a sense can be seen as reflecting aspects of both Computational Complications and Logical Lingerings as discussed before.
In \cite{DEALMEIDABORGES2020104636} it is shown that really huge formulas are needed to just express Article 8.6 in what is called Linear Temporal Logic. We refer the reader to the paper for details and just observe here that the numbers are not as big as 
those mentioned in Section \ref{section:ComputationalComplications} but large enough so to become unreadable for all practical purposes. On the other hand, Article 8.6 has the non-locality property as described above: legality in a week may depend on driving behaviour arbitrarily far away. It is essentially this dependence on something arbitrarily far away that makes the Halting problem undecidable: you never know if some time far away your program will halt and this time may be arbitrarily far away (or never come). Of course, the difference here is that the natural numbers extend towards the infinite whereas the driver file will always be finite and bounded by the lifetime of the driver. Therefore, we do not speak of genuine Logical Lingerings but only shades thereof.

\section{Good administration and the efficiency of inefficiency}

The principles of \emph{Good Governance} and of \emph{Good Administration} are lately much discussed \cite{Ponce:2022:Nudging}. They entail a general good and fair application of among others public administrations to citizens.

\subsection{Buono governo and pragmatics}

Apart from a guiding principle the principles of Good Governance and Good Administrations are actually also a right as anchored in various legal bodies \cite{GoodGovernanceRights}. The principle lends it name from a famous painting from Ambrogio Lorenzetti (1290 -- 1348) who is an important exponent of the so-called Sienese school of painters: \emph{The Allegory of Good and Bad Governance}. 

\begin{figure}[H]
 \begin{center}
  \includegraphics[scale=.71]{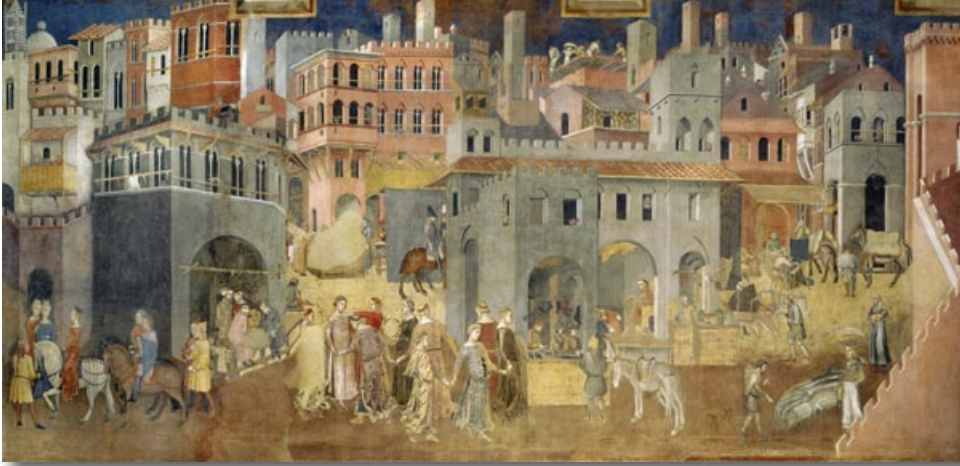}
\caption{The Allegory of Good and Bad Governance (Ambrogio Lorenzetti)}\label{figure:Lorenzetti}
 \end{center}
 \end{figure}
The name of the painting, the old zeal surrounding it and the context of the welfare of Siena tends to invoke an image of wise governors and administrators that guide themselves by justice and reason alone. In this context it is nice to mention one of the rules in place during the Sienese peak of wealth and rule: The army commander (\emph{Capitano}) was not allowed to live at a distance of less than 60 km from the Sienese city center. The rule was put in place for a pragmatic reason. The more important families lived in or near the city center where they could exercise their influence and power to the general but also their own benefit. If one of those families moreover had power over the army forces, this was deemed too high a risk for sole rule and tyranny. 

Behold good governance. In principle, the Sienese rule bears some injustice to excellent strategists and army commanders who happen to live in the city. Pragmatics, experience and knowledge of human nature however put this simple but effective rule in place. And this simple rule yielded longevity to the Sienese society. The rule can be seen as an example of the efficiency of inefficiency: local talent from the city center may be left unyielded by the rule which is inefficient but on the long run the rule avoids internal power clashes so that it is efficient indeed.

Nature uses natural selection so that self-enforcing rules abide. A beautiful example are fungus-growing termites. The termites are minute yet some species are able to produce mounds in the order of thousand times the size of an individual. The mounds are complex structures with ventilation shafts and the mounds are oriented in such a way to optimise sun exposure so that it creates close to constant temperature and humidity cavities inside the mounds where fungi can grow so that the fungi can produce a by-product that is essential to the termites. Though much is unknown about the exact organisation, it is likely that various relatively simple rules\footnote{Recall how we earlier observerd how simple rules give rise to complex behaviour like undecidability in the Halting problem or fractal structures in the Mandelbrot set.} underly the emerging of such rich and well functioning structures.

\subsection{Patients and health care}

Human society is a highly complex environment. Justice, logic and in general, good administration are leading principles in shaping our society. Somehow, it seems, these principles alone are not enough. Concrete rules that are to implement good administration often have side-effects that seem to overrule the founding principles of good administration. It is tempting to link simple mechanisms to observables in societies: those close to where decisions are being made about the height of wages having higher wages; construction rules yielding often uneasthetic buildings without proper thermic isolations where seemingly only the project developer is winning; those who shape the rule-based organisation of society seeing their existence reaffirmed by having many rules around thereby leading to a \emph{rulification} of our society; etc.

Leading principles are necessary and so are close monitoring of the effects of the rules that should implement these leading principles. The latter, as we have seen (recall Sections \ref{section:MultipleMathMischief}, \ref{section:ComputationalComplications}, \ref{section:LogicalLingerings}) is a highly complicated task if not sometimes impossible. A possible new leading principle could be the \emph{Efficiency of inefficiency} that allows considering simple rules that at first sight seem to go straight against the general purpose. We have seen the Sienese talent waste in the previous subsection. To illustrate the principle we like to mention a case from Spanish and Catalan Health care. 

General medical practice in Spain and Catalonia is offered through Health centers (called CAP in Catalonia) whose size depends on the municipality but that is typically around the ten general practitioners. Each CAP has a local coordinator and various CAPs are clustered into various organisational superstructures that eventually resonate at the national government level. As in our examples above, there is a tension. 

\begin{itemize}
\item[(A)]
On the one hand, coordinators should be good coordinators and familiar with the surrounding organising culture. As such, the system benefits from administrators with much managerial experience. 

\item[(B)]
On the other hand, it is important that those who occupy managerial positions, directly tap into the dynamics of the work floor. As such, the system benefits from administrators who recently really practiced the medical profession. 
\end{itemize}

Even though the Catalan health care is aware of these tensions, it seems that various mechanisms benefit (A) rather than (B). For example, a medical team coordinator has a minimal period in which she can serve but there is no maximal period stipulated. Work pressure is very high in the medical practice (especially after COVID with many practitioners with stress, illness or burn-out syndrome). Work pressure is also high in the medical managerial spheres but arguably less so or at least of a different nature in the higher strata of management. Managerial responsibilities come with salary stimuli starting with a gross increase of around 15 -- 20 \%. 

There are also mechanisms to favour (B) and more so in the lower strata of management. For example, the coordinator of a health care center needs to be a medical doctor or a nurse\footnote{Recently one also starts considering odontologists on occasion.}. 

It seems indeed that (A) is often winning from (B) so that professionals going into managerial positions stay there and lose contact with the work-floor, serve the need of fluent management over the need of patients and professionals. It is hard to decide on cause-and-effect but the facts are on the table with high tensions in the field and occupation of medical staff often at a bare minumum\footnote{We mention the Cardedeu center in particular where in fall 2023 only 4.5 GPs were at work where there should be 12 while at the top management of the Generalitat yearly salarial costs include around 13 million euros over 141 top managers. }.  

The efficiency of inefficiency could suggest a consideration here. On the short run it is clearly inefficient to change medical managers often so that new managers need to learn basic management. On the longer run this may however cause a dynamics that lowers the gap between patients and medical work-floor on the one hand, and (higher) management on the other hand. 

\subsection{Conclusions and the dreams we wish to dream}

In this paper we have studied AI in public administrations. Our evidence supported hypothesis was that rule-based AI is the only current form of AI that allows for full control and `understanding'. However, we have seen that even in rule-based IA various complications arise. We organised the various complications by their nature: logical, computational and mathematical and saw various examples, mostly from traffic regulations. We observed that leading principles are needed yet need to be accompanied by close monitoring of the rules in force. 

AI is an imprecise denominator of a wide field with fast development. As such it may seem wise to have regulations follow suit on a case-by-case fashion rather than following general guiding principles about AI that are to shape a field we are only starting to see the beginnings of. It is often mentioned that AI makes us reflect on the human condition and who we are as humans. We think that it is an equally important question to ask \emph{who we want to be} since AI seems to be providing us with a tool to shape our future. Again, here the \emph{efficiency of inefficiency} seems to play up. Sure enough, AI can speed up processes and perform a myriad of tasks. But do we really want these tasks to be automated? 

It is a good thing that AI can control diabetes. But, is it a good thing that human helpdesks are disappearing? Is it a good thing that elderly people have voice-connected chatbots or holograms to mitigate their solitude? Short-term efficiency seems to be affecting the core values of compassion and humanity. Is optimising short term efficiency really the dream we should be dreaming? Not all that can be done with AI to optimise short-term efficiency is actually improving the quality of our lives. Imagine calling a friend with whom you consider going for a drink. The answering machine of your friend recognises your number and addresses you by the name and offers you a choice menu: If you want to talk politics, press or say 1; If you want to meet for a chess-game, press or say 2; etc. It seems a healthy and natural reaction to simply hang up on this person and rather call some other friend. 

\section*{Acknowledgements} 
This paper is largely based on a keynote presentation by the first author in the conference \emph{Public Administration and the EU Proposal for a Regulation of Artificial Intelligence} so first of all we want to thank the organisers of that conference and in particular Agustí Cerrillo and Juli Ponce. Further, we would like to mention and thank colleagues for feedback, discussions and other contributions: Petia Guintchev, Moritz Müller,  Vicent Navarro, Cosimo Perini and, Martin Soto.

\bibliographystyle{plain}
\bibliography{References}

\begin{thebibliography}{10}

\bibitem{LeapSecondList}
leap-seconds.list, 2024.

\bibitem{BenDavid:2019:LearnabilityUndecidable}
Shai Ben-David, Pavel Hrube\u{s}, Shay Moran, and Amir Yehudayoff.
\newblock Learnability can be undecidable.
\newblock {\em Nature Machine Intelligence}, 1:44--49, 2019.

\bibitem{Boring:Intelligence:1923}
Edwin~G. Boring.
\newblock Intelligence as the tests test it.
\newblock {\em New Republic}, 36:35--37, 1923.

\bibitem{Corsi:2021:FVforNueralNetworks}
Davide Corsi, Enrico Marchesini, and Alessandro Farinelli.
\newblock Formal verification of neural networks for safety-critical tasks in
  deep reinforcement learning.
\newblock In Cassio de~Campos and Marloes~H. Maathuis, editors, {\em
  Proceedings of the Thirty-Seventh Conference on Uncertainty in Artificial
  Intelligence}, volume 161 of {\em Proceedings of Machine Learning Research},
  pages 333--343. PMLR, 27--30 Jul 2021.

\bibitem{GoodGovernanceRights}
{Council of European Union}.
\newblock Council regulation ({EU}) no 2012/c 326/1, 2012.

\bibitem{Davis:1958:ComputabilityUnsolvability}
M.~D. Davis.
\newblock {\em Computability and Unsolvability}.
\newblock McGraw-Hill, 1958.

\bibitem{DEALMEIDABORGES2020104636}
Ana {de Almeida Borges}, Juan~José {Conejero Rodríguez}, David
  Fernández-Duque, Mireia {González Bedmar}, and Joost~J. Joosten.
\newblock To drive or not to drive: A logical and computational analysis of
  european transport regulations.
\newblock {\em Information and Computation}, page 104636, 2020.

\bibitem{AlmeidaBorges:2022:PosterAchtProcent}
Ana de~Almeida~Borges, Mireia Gonz\'{a}lez~Bedmar, Juan
  Conejero~Rodr\'{\i}guez, David Fernández~Duque, and Joost~J. Joosten.
\newblock To drive or not to drive: A logical and computational analysis of
  european transport regulations, 2022.

\bibitem{AlmeidaBorges:2024:TimeLibrary}
Ana de~Almeida~Borges, Mireia Gonz\'{a}lez~Bedmar, Juan
  Conejero~Rodr\'{\i}guez, Eduardo Hermo~Reyes, Joaquim Casals Bu\~{n}uel, and
  Joost Joosten.
\newblock Utc time, formally verified.
\newblock CPP 2024, page 2–13, New York, NY, USA, 2024. Association for
  Computing Machinery.

\bibitem{EWD898}
Edsger Dijkstra.
\newblock The threats to computing science.

\bibitem{Errezil:2019:Homologation}
G.~Errezil~Alberdi.
\newblock Industrial {S}oftware {H}omologation: {T}heory and case study.
  {I}ndustrial {S}oftware {H}omologation: {T}heory and case study {A}nalysis of
  the {E}uropean tachograph technology with {EU} transport {R}egulations
  3821/85, 799/2016, and 561/06 and their consequences for {E}uropeans
  citizens.
\newblock Technical report, Formal Vindications S.L., 2019.

\bibitem{Regulation799}
{European Parliament and Council of the European Union}.
\newblock Commission implementing {R}egulation ({EU}) 2016/799 of 18 {M}arch
  2016 implementing {R}egulation ({EU}) {N}o 165/2014 of the {E}uropean
  {P}arliament and of the {C}ouncil laying down the requirements for the
  construction, testing, installation, operation and repair of tachographs and
  their components.
\newblock Official Journal of the European Union, 2016.

\bibitem{FernandezDuqueEtAl:2019:DriveOrNotRegulations50Etc}
David Fernández-Duque, Mireia {González Bedmar}, {Daniel Sousa}, {Joost J.
  Joosten}, and {Guillermo Errezil Alberdi}.
\newblock To drive or not to drive: A formal analysis of {R}equirements (51)
  and (52) from {R}egulation ({EU}) 2016/799.
\newblock In {\em Personalidades jur{\'{\i}}dicas difusas y artificiales},
  TransJus Working Papers, pages 159--171. Institut de Recerca TransJus, 2019.

\bibitem{FultonPlatzer:2018:SafeLearning}
Nathan Fulton and André Platzer.
\newblock Safe reinforcement learning via formal methods: Toward safe control
  through proof and learning.
\newblock {\em Proceedings of the AAAI Conference on Artificial Intelligence},
  32(1), 2018.

\bibitem{Guintchev:2024:LanguageVsPrinciples}
P~Guintchev, J.~J. Joosten, S.~Santiago~Fernández, E.~Sancho~Adamson, and
  M.~Soria~Heredia.
\newblock Specification languages for computable laws versus basic legal
  principles (to appear), 2024.

\bibitem{Collatz}
Richard~K. Guy.
\newblock {\em E16: The 3x+1 problem}, pages 330--336.
\newblock Springer New York, 3 edition.

\bibitem{TaymiyyaAgainstGreeks}
Wael B.~(ed.) Hallaq.
\newblock {xiiIntroduction}.
\newblock In {\em {Ibn Taymiyya Against the Greek Logicians}}. Oxford
  University Press, 09 1993.

\bibitem{Hutson:2024:TwoFacedAI}
Matthew Hutson.
\newblock Two-faced ai language models learn to hide deception.
\newblock {\em Nature}, 2024.

\bibitem{HuttnerMerigoux:2022:CatalaFuture}
Liane Huttner and Denis Merigoux.
\newblock {Catala: Moving Towards the Future of Legal Expert Systems}.
\newblock {\em {Artificial Intelligence and Law}}, August 2022.

\bibitem{UTCStandard}
{ITU Radiocommunication Assembly}.
\newblock Recommendation itu-r tf.460-6: Standard-frequency and time-signal
  emissions.
\newblock {\em International Telecommunication Union}, 2002.

\bibitem{JespersenCarara:Malfunction:2011}
Bj\o{}rn Jespersen and Massimiliano Carrara.
\newblock Two conceptions of technical malfunction.
\newblock {\em Theoria}, 77(2):117--138, 2011.

\bibitem{Joosten:2023:Dialogues}
Joost~J. Joosten.
\newblock Dialogues with algorithms, 2023.

\bibitem{Kleene:1952:IntroductionMetamathematics}
S.C. Kleene.
\newblock {\em Introduction to Metamathematics}.
\newblock Wolters-Noordhof and North-Holland, 1952.

\bibitem{merigoux:2023:Rules}
Denis Merigoux, Marie Alauzen, and Lilya Slimani.
\newblock {Rules, Computation and Politics: Scrutinizing Unnoticed Programming
  Choices in French Housing Benefits}.
\newblock {\em Journal of Cross-disciplinary Research in Computational Law},
  1(3), 2023.
\newblock (forthcoming).

\bibitem{MerigouxEtAl:2021:Catala}
Denis Merigoux, Nicolas Chataing, and Jonathan Protzenko.
\newblock Catala: A programming language for the law.
\newblock {\em Proceedings of the ACM on Programming Languages}, 5, 2021.

\bibitem{MonatMerigoux:2024:DatesAmbiguitiesLaw}
Raphaël Monat, Aymeric Fromherz, and Denis Merigoux.
\newblock Formalizing date arithmetic and statically detecting ambiguities for
  the law.
\newblock In {\em Programming Languages and Systems}, ESOP, Cham, 2024.
  Springer Nature Switzerland.
\newblock forthcoming.

\bibitem{mullerJoosten:2023:modelcheckingPreprint}
Moritz Müller and Joost~J. Joosten.
\newblock Model-checking in the foundations of algorithmic law and the case of
  regulation 561.
\newblock {\em ArXiv:2307.05658 [cs.LO]}, 2023.

\bibitem{Oregan:2017:GuideFormalMethods}
G.~O'Regan.
\newblock {\em Concise Guide to Formal Methods: Theory, Fundamentals and
  Industry Applications}.
\newblock Undergraduate Topics in Computer Science. Springer International
  Publishing, 2017.

\bibitem{Papadimitriou:1994:ComputationalComplexity}
C.H. Papadimitriou.
\newblock {\em Computational Complexity}.
\newblock Theoretical computer science. Addison-Wesley, 1994.

\bibitem{Reg561}
European Parliament and Council of~the European~Union.
\newblock Regulation (ec) no 561/2006 of the {E}uropean {P}arliament and of the
  {C}ouncil of 15 march 2006 on the harmonisation of certain social legislation
  relating to road transport.
\newblock {\em Official Journal of the European Union}, 2006.

\bibitem{Ponce:2022:Nudging}
J.~(Ed.) Ponce.
\newblock {\em Nudging’s Contributions to Good Governance and Good
  Administration – Legal Nudges in Public and Private Sectors}.
\newblock European Public Law Organization, 2022.

\bibitem{Roivainen:IQChatGPT:2023}
Eka Roivainen.
\newblock I gave chatgpt an iq test. here’s what i discovered, 2023.

\bibitem{RussellNorvig:AIAmodernApproach:2010}
Stuart Russell and Peter Norvig.
\newblock {\em Artificial Intelligence: A Modern Approach}.
\newblock Pearson, 3 edition, 2010.

\bibitem{UnderfittingUndecidable}
Sonia Sehra, David Flores, and George~D. Montañez.
\newblock Undecidability of underfitting in learning algorithms.
\newblock In {\em 2021 2nd International Conference on Computing and Data
  Science (CDS)}, pages 591--594, 2021.

\bibitem{Turing:1936:ComputableNumbers}
A.~M. Turing.
\newblock On computable numbers, with an application to the entscheidungs
  problem.
\newblock {\em Proceedings of the London Mathematical Society}, 2(1):230–265,
  1936.

\bibitem{Turing:1950:Intelligence}
A.~M. Turing.
\newblock Computing machinery and intelligence.
\newblock {\em Mind}, 59(236):433--460, 1950.

\bibitem{Wason:ReasoningAboutARule:1968}
P.~C. Wason.
\newblock Reasoning about a rule.
\newblock {\em Quarterly Journal of Experimental Psychology}, 20(3):273--281,
  1968.

\bibitem{WasonShapiro:Reasoning:1971}
P.~C. Wason and Diana Shapiro.
\newblock Natural and contrived experience in a reasoning problem.
\newblock {\em Quarterly Journal of Experimental Psychology}, 23(1):63--71,
  1971.

\bibitem{Wegener:2005:ComplexityTheory}
Ingo Wegener.
\newblock {\em Complexity Theory}.
\newblock Springer Berlin, Heidelberg, 2005.

\bibitem{Wiklicky:1994:LoadingProblemUndecidable}
H.~Wiklicky.
\newblock The neural network loading problem is undecidable.
\newblock {\em Proceedings of the First European Conference on Computational
  Learning Theory}, pages 183--192, 1994.

\end{thebibliography}

\end{document}